%% file: main.tex
\documentclass[lettersize,journal]{IEEEtran}
\usepackage{amsmath,amsfonts}
\usepackage{algorithmic}
\usepackage{algorithm}
\usepackage{array}
\usepackage[caption=false,font=normalsize,labelfont=sf,textfont=sf]{subfig}
\usepackage{textcomp}
\usepackage{stfloats}
\usepackage{url}
\usepackage{verbatim}
\usepackage{graphicx}
\usepackage{cite}
\usepackage{booktabs}
\usepackage{colortbl}
\usepackage{multirow}
\usepackage[table,xcdraw]{xcolor}
\usepackage{soul}
\usepackage{subfig}

\definecolor{lightgray}{gray}{0.9}

\usepackage{hyperref}
\hypersetup{
    colorlinks=true,
    linkcolor=blue,
    filecolor=blue,      
    urlcolor=blue,
    citecolor=blue,
    pdftitle={CHAD},
    }

\begin{document}

\title{VegaEdge: Edge AI Confluence Anomaly Detection for Real-Time Highway IoT-Applications}

\author{Vinit Katariya,~\IEEEmembership{Student Member,~IEEE,}
        Fatema-E- Jannat,~\IEEEmembership{Student Member,~IEEE,}
        Armin Danesh Pazho\IEEEauthorrefmark{1},~\IEEEmembership{Student Member,~IEEE,}
        Ghazal Alinezhad Noghre,~\IEEEmembership{Student Member,~IEEE,}
        Hamed Tabkhi,~\IEEEmembership{Member,~IEEE}
\thanks{The authors are with the Electrical and Computer Engineering Department, The University of North Carolina at Charlotte, Charlotte,
	NC, 28223 USA.\\
	\{vkatariy, fjannat, adaneshp, galinezh, htabkhiv\}@uncc.edu\\\IEEEauthorrefmark{1} Corresponding author.
	}}

\maketitle

\begin{abstract}
Vehicle anomaly detection plays a vital role in highway safety applications such as accident prevention, rapid response, traffic flow optimization, and work zone safety. With the surge of the Internet of Things (IoT) in recent years, there has arisen a pressing demand for Artificial Intelligence (AI) based anomaly detection methods designed to meet the requirements of IoT devices. Catering to this futuristic vision, we introduce a lightweight approach to vehicle anomaly detection by utilizing the power of trajectory prediction. Our proposed design identifies vehicles deviating from expected paths, indicating highway risks from different camera-viewing angles from real-world highway datasets. On top of that, we present \emph{VegaEdge} – a sophisticated AI confluence designed for real-time  security and surveillance applications in modern highway settings through edge-centric IoT-embedded platforms equipped with our anomaly detection approach. Extensive testing across multiple platforms and traffic scenarios showcases the versatility and effectiveness of VegaEdge. This work also presents the \emph{Carolinas Anomaly Dataset (CAD)}, to bridge the existing gap in datasets tailored for highway anomalies. In real-world scenarios, our anomaly detection approach achieves an AUC-ROC of 0.94, and our proposed VegaEdge design, on an embedded IoT platform, processes 738 trajectories per second in a typical highway setting. The dataset is available at \url{https://github.com/TeCSAR-UNCC/Carolinas_Dataset#chd-anomaly-test-set}.

\end{abstract}


\input{tex/introduction}

\input{tex/related_works}

\input{tex/Anomaly_Detection_Dataset}

\input{tex/Anomaly_Detection_Algorithm}

\input{tex/design}

\input{tex/results}

\section{Conclusion}
In this work introduced a minimalist anomaly detection approach based on predicted trajectories and showcased its effectiveness across various prediction windows on adversarial and real-world anomalies. We presented VegaEdge, a real-time highway safety solution optimized IoT-edge applications utilizing our anomaly detection method, achieving a throughput of up to 758 processed vehicle trajectories per second in high-traffic conditions. Furthermore, our application of VegaEdge demonstrated its ability to adapt buffer times for workzone personnel, highlighting the trade-off between buffer time and accuracy for such applications. The introduction of the Carolinas Anomaly Dataset (CAD) as a dedicated resource for real-world highway anomaly detection, combined with our innovative approach, highlights the potential of IoT and AI in advancing highway safety.

\section*{Acknowledgments}
This research is supported by the National Science Foundation (NSF) under Award No. 1932524.

\bibliographystyle{IEEEtran}
\bibliography{bibliography}

\section{Biography Section}
 
\vspace{-35pt}
\begin{IEEEbiography}[{\includegraphics[width=1in,height=1.25in,clip,keepaspectratio]{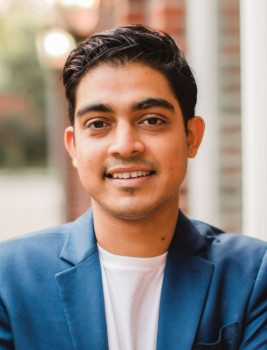}}]
{Vinit Katariya} (S'14) is a PhD Candidate at the Department of Electrical and Computer Engineering, University of North Carolina at Charlotte, USA. He received his Master's degree from the University of North Carolina at Charlotte in 2016. His work addresses challenges in the fields of Computer Vision, Deep Learning, and Intelligent Transportation. His current research focuses on real-time Artificial Intelligence frameworks and applications for real-world systems utilizing embedded-edge platforms.
\end{IEEEbiography}

\vspace{-35pt}
\begin{IEEEbiography}
[{\includegraphics[width=1in,height=1.12in,keepaspectratio]{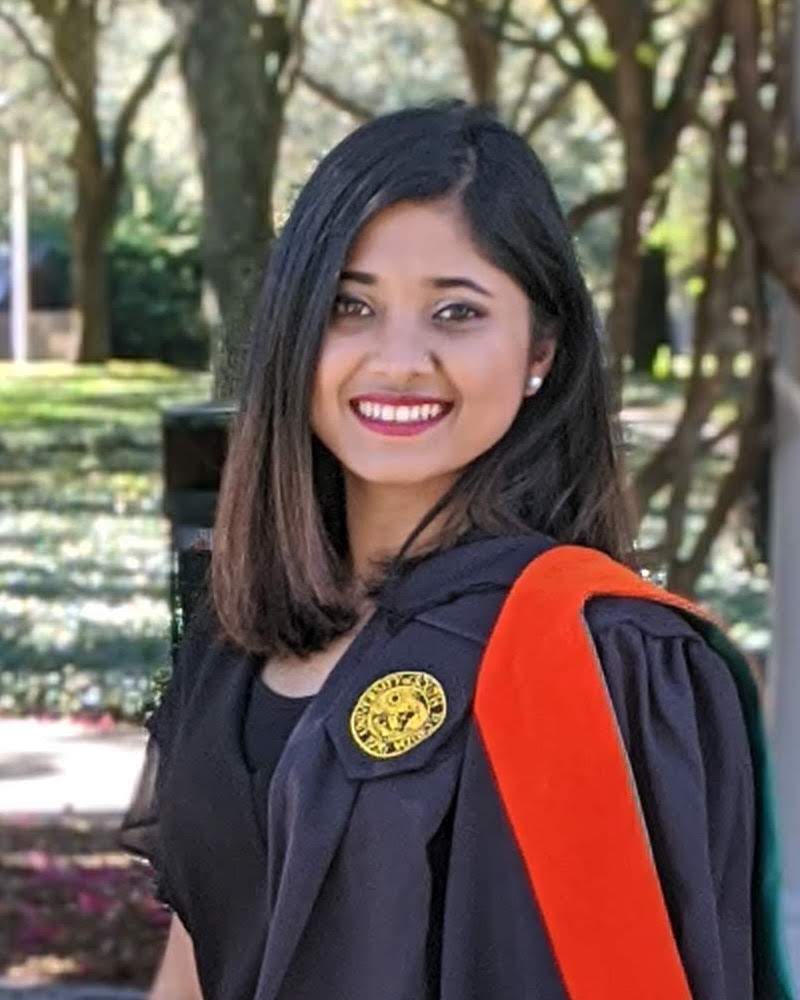}}]
{Fatema-E- Jannat} (S’22) is currently pursuing her Ph.D. in Electrical and Computer Engineering at the University of North Carolina at Charlotte, USA. She holds a Master's degree from the University of South Florida, USA, earned in 2019. Her research focuses on the strategic application of deep-learning techniques specifically in the domain of object detection, anomaly detection, and intelligent transportation systems aiming to improve societal well-being.  
\end{IEEEbiography}

\vspace{-35pt}
\begin{IEEEbiography}[{\includegraphics[width=1in,height=1.12in,keepaspectratio]{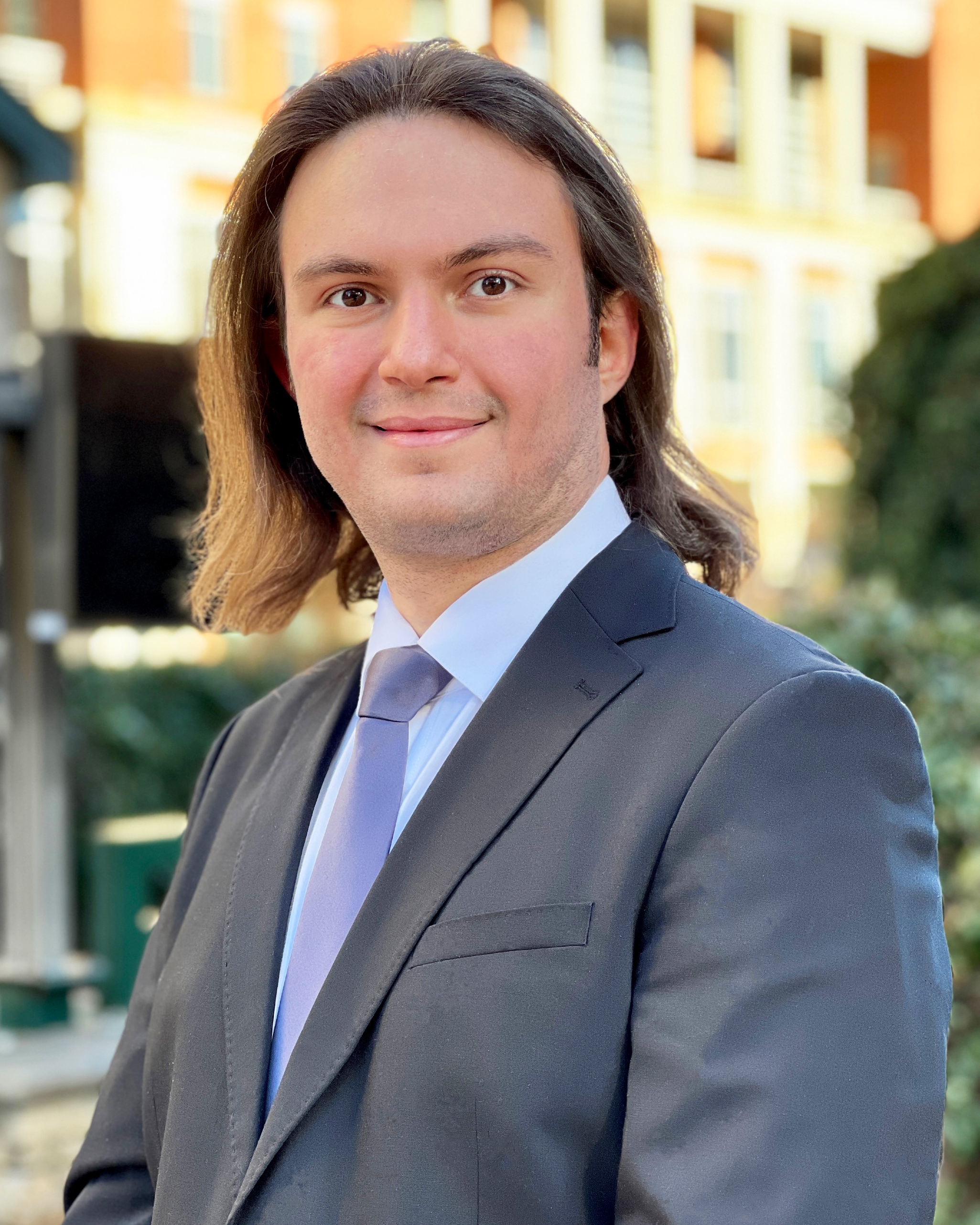}}]{Armin Danesh Pazho} (S’22) is currently a Ph.D. student at the University of North Carolina at Charlotte, NC, United States. With a focus on Artificial Intelligence, Computer Vision, and Deep Learning, his research delves into the realm of developing AI for practical, real-world applications and addressing the challenges and requirements inherent in these fields. Specifically, his research covers action recognition, anomaly detection, person re-identification, human pose estimation, and path prediction.
\end{IEEEbiography}

\vspace{-35pt}
\begin{IEEEbiography}[{\includegraphics[width=1in,height=1.1in,keepaspectratio]{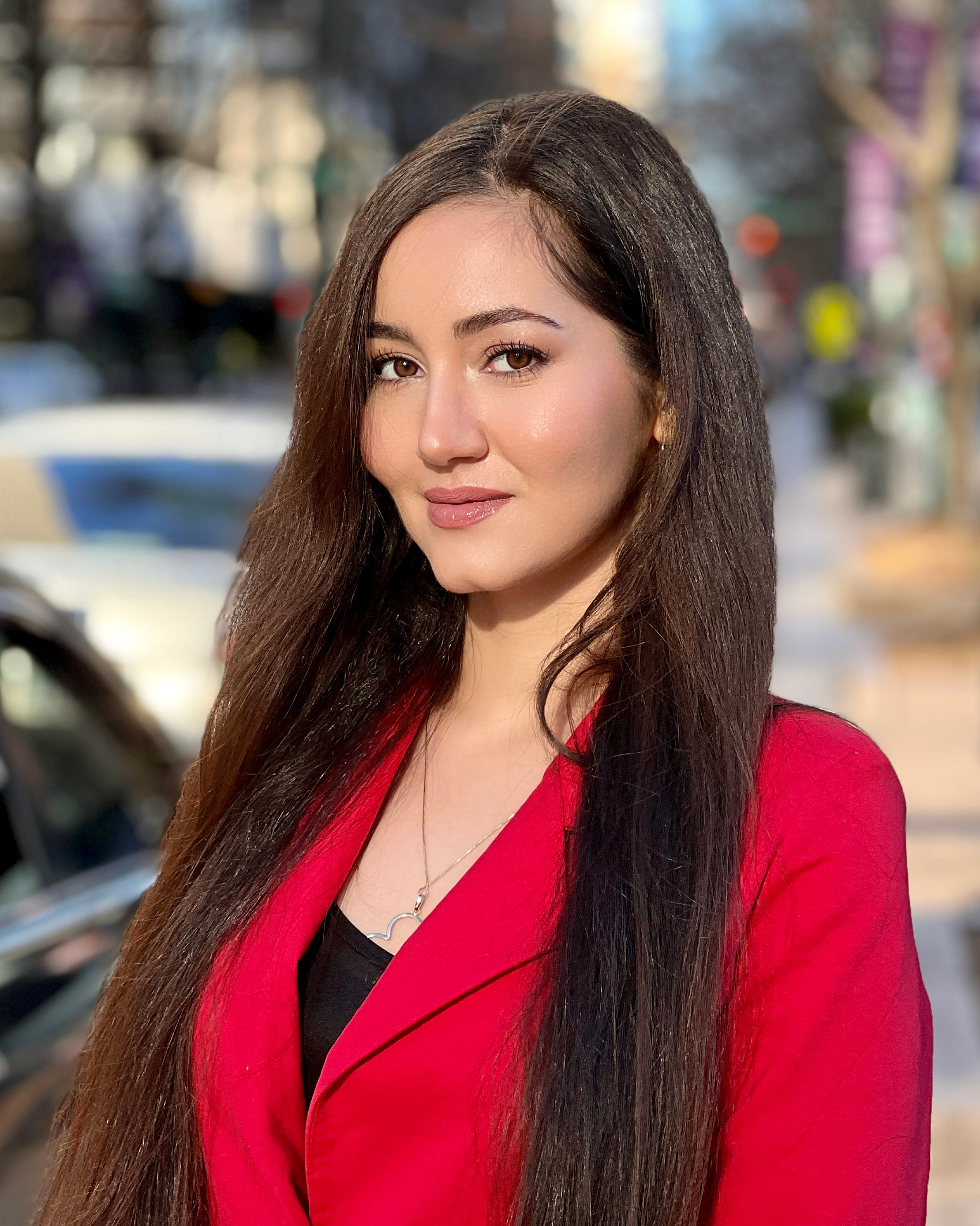}}]{Ghazal Alinezhad Noghre} (S’22) is currently pursuing her Ph.D. in Electrical and Computer Engineering at the University of North Carolina at Charlotte, NC, United States. Her research concentrates on Artificial Intelligence, Machine Learning, and Computer Vision. She is particularly interested in the applications of anomaly detection, action recognition, and path prediction in real-world environments, and the challenges associated with these fields.
\end{IEEEbiography}

\vspace{-35pt}
\begin{IEEEbiography}[{\includegraphics[width=1in,height=1.25in,keepaspectratio]{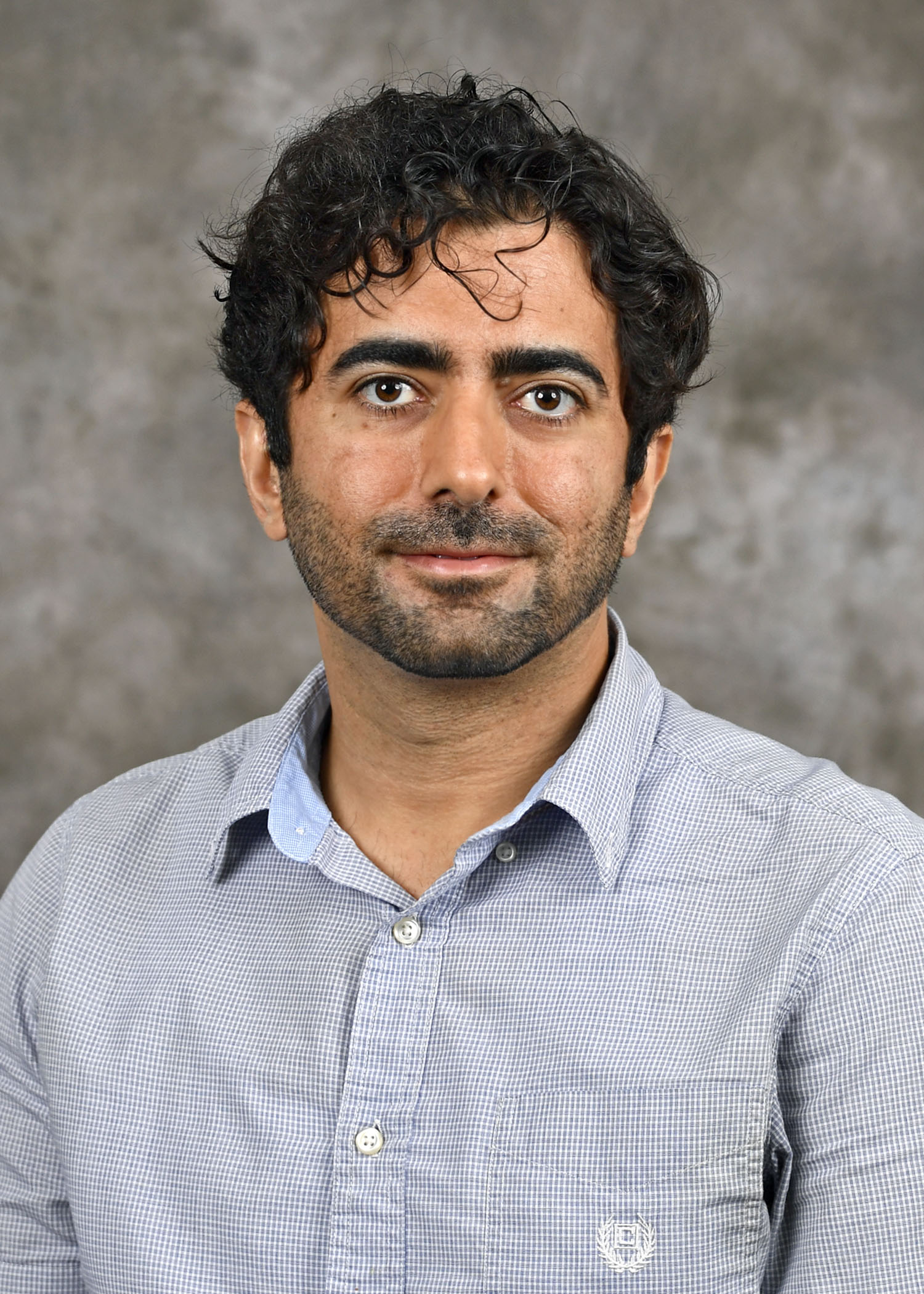}}]{Hamed Tabkhi}
(S’07–M’14)
is an Associate Professor in the Department of Electrical and Computer Engineering, University of North Carolina at Charlotte, USA.
He was a post-doctoral research associate at Northeastern University. Hamed Tabkhi received his Ph.D. in 2014 from Northeastern University under the direction of Prof. Gunar Schirner. His research focuses on transformative computer systems and architecture for cyber-physical, real-time streaming, and emerging machine learning applications.
\end{IEEEbiography}

\vfill

\end{document}

%% file: tex/introduction.tex
\section{Introduction}
In today's digital age dominated by the Internet of Things (IoT), camera-based infrastructure has become an integral part of our interconnected world. With urbanization intensifying, our highways face increasing congestion and unpredictable driving patterns. Although current highway cameras offer surveillance, their true potential to harness real-time analytics remains largely untapped. Integrating edge-based AI frameworks with these cameras can revolutionize traffic management and safety \cite{sabeti2021toward}. This integration not only promises rapid detection and response to anomalies but also increases bandwidth efficiency, lowers latency, and scales highway monitoring, marking a transformative approach to road safety and management.

\begin{figure}[t]    
    \centering
    \resizebox{1\columnwidth}{!}{
    \includegraphics[clip,trim={18 23 18 18},width=1\columnwidth]{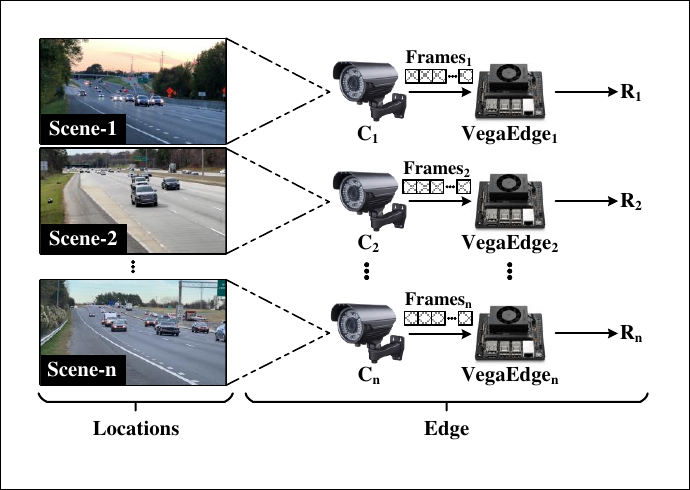}
    }
    \caption{Example demonstration of VegaEdge implementation on an IoT platform for real-world highway scenarios.}
    \label{fig:intro}
\end{figure}

The AI-based edge applications can help with real-time detection of erratic driving behaviors that can help tackle the distressing surge in accidents, especially within work zones. From 2003 to 2020, worker fatalities rose, with 135 deaths in 2019 and 117 in 2020 \cite{nnaji2018work, fhwaPublication2022_wz_sign}. The Federal Highway Administration's 2021 report highlighted 106,000 work zone accidents, resulting in 42,000 injuries and 956 fatalities \cite{nhtsa2022, fhwa2021_facts_stats}. While traditional safety mechanisms in these zones are primarily reactive \cite{nnaji2020improving}, often leading to late interventions, integrating AI at the data's edge ensures timely decision-making crucial for highway safety, surveillance, and traffic analysis applications.

Anomaly detection for roadways mainly focuses on the complexities of autonomous driving \cite{hu2023detecting} in urban settings, where interactions among vehicles, infrastructure, and pedestrians are intricate. Influenced by factors like intersections and diverse road alignments, urban trajectories are notably unpredictable. In contrast, highway travel, designed for longer distances, exhibits more predictable behaviors \cite{tian2023earth, izquierdo2023vehicle, radermecker2023, Chen2023lane_change}. 

For effective AI in highway settings, models need training on specific datasets differentiating normal from abnormal driving. Many existing datasets, however, lack resolution and relevance. Addressing this, we present the \emph{Carolinas Anomaly Dataset (CAD)} with real-world highway anomalies. Moreover, real-time anomaly detection is crucial for edge-based safety applications prioritizing nimbleness. CAD emphasizes identifying vehicle trajectories that deviate from standard paths, especially those moving outside of their lanes. While many anomaly detection methods exist \cite{deng2021anomaly,wu2021box,li2020multi}, they're not highway-specific and require significant computational power. Overall, there is a clear deficiency in real-world highway anomaly datasets and corresponding algorithms, leading to a void in AI frameworks for real-time anomaly detection in practical applications. 

In this context, we introduce \emph{VegaEdge}, an edge AI confluence tailored for real-time highway IoT applications operating on embedded edge devices using lightweight anomaly detection. Fig. \ref{fig:intro} shows how VegaEdge can monitor highway traffic and detect anomalies at the edge at various locations on embedded platforms. Our evaluations on real-world datasets confirm VegaEdge's effectiveness and its performance with real-world video detection will be detailed in subsequent sections.
In this paper, we also introduce a lightweight method that uses ground truth and trajectory prediction for quick anomaly detection, promoting enhanced highway safety. This method seamlessly integrates with the State-of-the-Art (SotA) trajectory prediction \cite{katariya2023pov} integrated with VegaEdge providing swift real-world anomaly detection.

We extensively evaluated VegaEdge across three distinct platforms, two of which are edge-based, low-power devices, underscoring its versatility. Tests were conducted to validate its robustness using real-world and simulated videos, demonstrating VegaEdge's capability to function with digital twins and across varying traffic densities. Furthermore, we examined its performance specifically for highway work zone safety, analyzing the impact of diverse prediction windows on the buffer times afforded to workers during potential hazards. The efficacy of our anomaly detection is showcased through evaluations on both adversarial and real-world datasets. These tests underscore the pronounced differences between adversarial-generated trajectories and real-world scenarios, emphasizing the imperative of employing real-world videos in authentic system deployments. We also perform extensive power analysis on an embedded to provide insights into the power consumption of VegaEdge in different power modes that can be utilized based on the desired application.

The main contributions of this paper are summarized as follows:

\begin{itemize}
\item We introduce Carolinas Anomaly Dataset (CAD), a new real-world anomaly dataset for highway applications. This dataset empowers researchers to validate anomaly detection techniques within genuine highway contexts.

\item We present a novel anomaly detection technique that foresees anomalous driving behaviors out of the predicted trajectories by extrapolating angel-based and displacement errors. Its effectiveness is demonstrated with adversarial and real-world trajectories on select datasets.

\item  We introduce VegaEdge, a cutting-edge AI-powered IoT solution for vehicle anomaly detection designed for edge-based embedded systems. It is adept at identifying vehicles that diverge from their anticipated route, indicating possible hazardous intrusions on highways in real-time.

\item  We subject VegaEdge and proposed anomaly detection techniques to exhaustive evaluations across multiple platforms and scenarios. The results showcase its adaptability and superior performance in real-world and simulated environments. We also demonstrate its effectiveness, emphasizing its application in work zone safety.

\end{itemize}

%% file: tex/related_works.tex
\section{Related Works}

Efforts have been made to adapt anomaly and vision models for IoT devices. \cite{ancilia_pazho} presents an IoT-focused video surveillance system, primarily analyzing human-related events. \cite{vision_iot} explores vision model applications in IoT, while \cite{anomaly_iot} investigates anomaly detection in time series data for domains like smart cities. Recently, there's been increased focus on highway safety. \cite{chandra2019robusttp} introduces a trajectory prediction framework for dense traffic, utilizing LSTMs and CNNs. \cite{liu2020vision} uses road geometry for vehicle counting, speed estimation, and classification. \cite{wei2019city} suggests a real-time flow estimation system based on pairwise scoring for vehicle counting. MultEYE \cite{balamuralidhar2021multeye} is an aerial viewpoint vehicle tracking system, leveraging segmentation for detection accuracy in edge devices and IoT applications. Nevertheless, a notable gap persists within AI-based solutions for highway applications, primarily due to the limited availability of real-world datasets and dedicated frameworks tailored specifically to highway-based edge applications with real-time processing capabilities. 

Anomaly detection in vehicle frameworks has been explored in various studies. \cite{deng2021anomaly} proposes an IoT system detecting abnormal driving using semantic analysis, vehicle detection, and 5G communication. \cite{peri2020towards} employs re-identification and multi-camera tracking with Gaussian Mixture Models (GMMs) to analyze vehicles. Anomalies are identified based on foreground-background changes. \cite{wu2021box} offers a tracking algorithm for anomaly detection in road scenes. \cite{tak2021development} enhances vehicle anomaly detection accuracy by integrating road geometry with movement predictions. \cite{li2020multi} presents a multi-granularity design combining various tracking levels for vehicle anomalies. Using clustering, \cite{jiang2009detecting} introduces a probabilistic framework for anomaly detection via vehicle trajectories.

Several studies, such as DSAB \cite{hu2023detecting}, focus on the vehicle anomaly detection problem individually. DSAB reconstructs vehicle social graphs using the Recurrent Graph Attention Network. \cite{jiao2023learning} employs Graph Convolutional Networks (GCNs) with a contrastive encoder for feature extraction, with the features later used in an SVM classifier. They also explore unsupervised methods using an Adversarial Autoencoder.

While there's a scarcity of comprehensive vehicle datasets in highway safety due to data gathering challenges, AI City Challenge offers a benchmark \cite{naphade20237th, naphade20192019}. Still, its alignment with highway safety is limited. The Carolinas Highway Dataset (CHD) \cite{katariya2023pov} provides videos from multiple viewpoints, ideal for highway safety. Given the rarity of anomalous driving behaviors, \cite{zhang2022adversarial} suggests an adversarial framework to generate anomalies on existing datasets. Recent studies like \cite{jiao2023learning} are utilizing this approach for more exhaustive anomaly detection evaluations. This lack of resources and use of adversarial approaches underscores the urgency of developing and advancing real-world datasets and AI-based IoT-edge solutions that capable of handling the unique challenges and anomalies of highway safety and surveillance.

%% file: tex/Anomaly_Detection_Dataset.tex
\vspace{-10pt}

\section{CAD: Carolinas Anomaly Dataset}
\label{sec:CAD}
Building upon discussions from prior sections, the absence of dedicated highway-based trajectory anomaly datasets presents a challenge in validating our anomaly detection methodologies. To circumvent this limitation, we adopt the recent advancements in adversarial anomaly generation as a testing bed for our proposed approach. Zhang et al. \cite{zhang2022adversarial}, introduce an adversarial attack-based technique designed to craft realistic anomalous trajectories by perturbing standard trajectories within a dataset.

While adversarial approaches have made significant strides in improving the fidelity of generated results, they still fall short of perfectly mirroring real-world scenarios. Although the disparity between machine-generated anomalies and actual real-world anomalies has diminished, it has not been completely eradicated. In light of the challenges inherent in evaluating highway anomalies and the existing gap in relevant datasets, we present the "Carolinas Anomaly Dataset (CAD)". This dataset, derived from the Carolinas Highway Dataset (CHD) \cite{katariya2023pov}, encompasses 22 videos, each exhibiting at least one anomalous driving trajectory. These videos are captured from two distinct vantage points: high-angle and eye-level, offering a versatile tool for surveillance and road safety applications. Specifically, CAD is composed of one-minute video segments, evenly split between the two perspectives, showcasing variety of anomaly behavior. 

In this context, anomalous behavior pertain to the atypical movement patterns exhibited by vehicles, including actions such as vehicles deviating from their designated lanes on the highway, abruptly halting in front of the camera's view, or vehicles that approach the camera while diverging away from their designated lane. These unusual and non-standard behaviors have the potential to pose significant risks to nearby structures, infrastructure, and, most critically, to the safety of workers, particularly within the dynamic and often high-speed environment of highway work zones.Designed to enable the evaluation of various anomaly detection methodologies, CAD serves as an invaluable resource for researchers focused on innovating highway safety through anomaly detection algorithms.

%% file: tex/Anomaly_Detection_Algorithm.tex
\section{Anomaly detection Methodology}
\label{sec:Anomaly_Detection}

In this section, we present our methods for anomaly detection using predicted trajectories. For anomaly detection, the trajectory prediction output is used to evaluate the error and angle-based approaches. The goal was to evaluate both methodologies for detecting anomalous behavior in trajectory and video datasets. Through this methodology, we allow the detection of unusual vehicle behaviors, such as sudden lane changes, erratic driving, or potential security threats in desired applications with minimum computation. 

\subsection{ADE-based Anomaly Detection}
\label{subsec:ADE-based}
Average Displacement Error (ADE) based Anomaly Detection is a method of computing the average error from the predicted trajectory to assess the accuracy of trajectory predictions for vehicles and comparing it against a threshold ($T^{ADE}_{anomaly}$) as per the application. It computes the average Euclidean distance between predicted trajectories ($\hat{F}$) and actual trajectories (${P}$) overall predicted time steps ($T_{pred}$) and subjects (${n}$) in a scene. The ADE in predicted trajectory from last $T_{past}$ seconds is compared with the desired ADE threshold, $T^{ADE}_{anomaly}$ as:

\begin{equation}
\frac{1}{n * T_{past}} \sum_{v=1}^{n} \sum_{t=1}^{T_{past}} \left| \hat{F}_{v}^{t} - {P}_{v}^{t} \right|_{2} > T^{ADE}_{anomaly},
\end{equation}

where ${t = 1, ..., T_{in}}$ is time and, ${v \in {1, 2, ..., n}}$ representing the index of the vehicle. By setting a threshold, a criterion is set to identify anomalous trajectories. Value exceeding the threshold indicates a significant disparity between the predicted and ground truth trajectories, resulting from unexpected driving behavior, going off the lane, etc. As PishguVe is designed to predict the normal trajectory of the ego vehicle, the ADE for any vehicle that exceeds the threshold is marked as an anomaly.

\subsection{Angle-based Anomaly Detection}
\label{subsec:angle-based}
Angle-based Anomaly Detection calculates the angle between the predicted future trajectory vector, \( \hat{F}_v \) and the actual trajectory vector, \( P_v \) of the ego vehicle and compares it with a threshold according to the application. Given the x and y coordinates of \( \hat{F}_v \) and \( P_v \)  for past few seconds $t_{past}$, we can compute the direction vectors:


\begin{equation}
\begin{array}{cccc}
D_{P_v} & = \begin{bmatrix} x_{v,p}^{t_{past}} - x_{v,p}^0 \\ y_{v,p}^{t_{past}} - y_{v,p}^0 \end{bmatrix} &
D_{\hat{F}_v} & = \begin{bmatrix} x_{v,f}^{t_{past}} - x_{v,f}^0 \\ y_{v,f}^{t_{past}} - y_{v,f}^0 \end{bmatrix}
\end{array}
\end{equation}

, here $x_{v,f}^{0}$ and $x_{v,f}^{0}$ are the position of vehicle a frame before the start of prediction. The angle between these direction vectors \( D_{P_v} \) and \( D_{\hat{F}_v} \) is compared with the threshold, $T^{Angle}_{anomaly}$ as:


\begin{equation}
\arccos \left( \frac{D_{P_v} \cdot D_{\hat{F}_v}}{\| D_{P_v} \| \| D_{\hat{F}_v} \|} \right) > T^{Angle}_{anomaly},
\end{equation}

where \( D_{P_v} \cdot D_{\hat{F}_v} \) is the dot product of \( D_{P_v} \) and \( D_{\hat{F}_v} \), and \( \| D_{P_v} \| \) and \( \| D_{\hat{F}_v} \| \) are their respective magnitudes.


%% file: tex/design.tex
\begin{figure*}[t]
    
    \centering
    \resizebox{1\linewidth}{!}{
    \includegraphics[clip,trim={18 35 46 18},width=1\columnwidth]{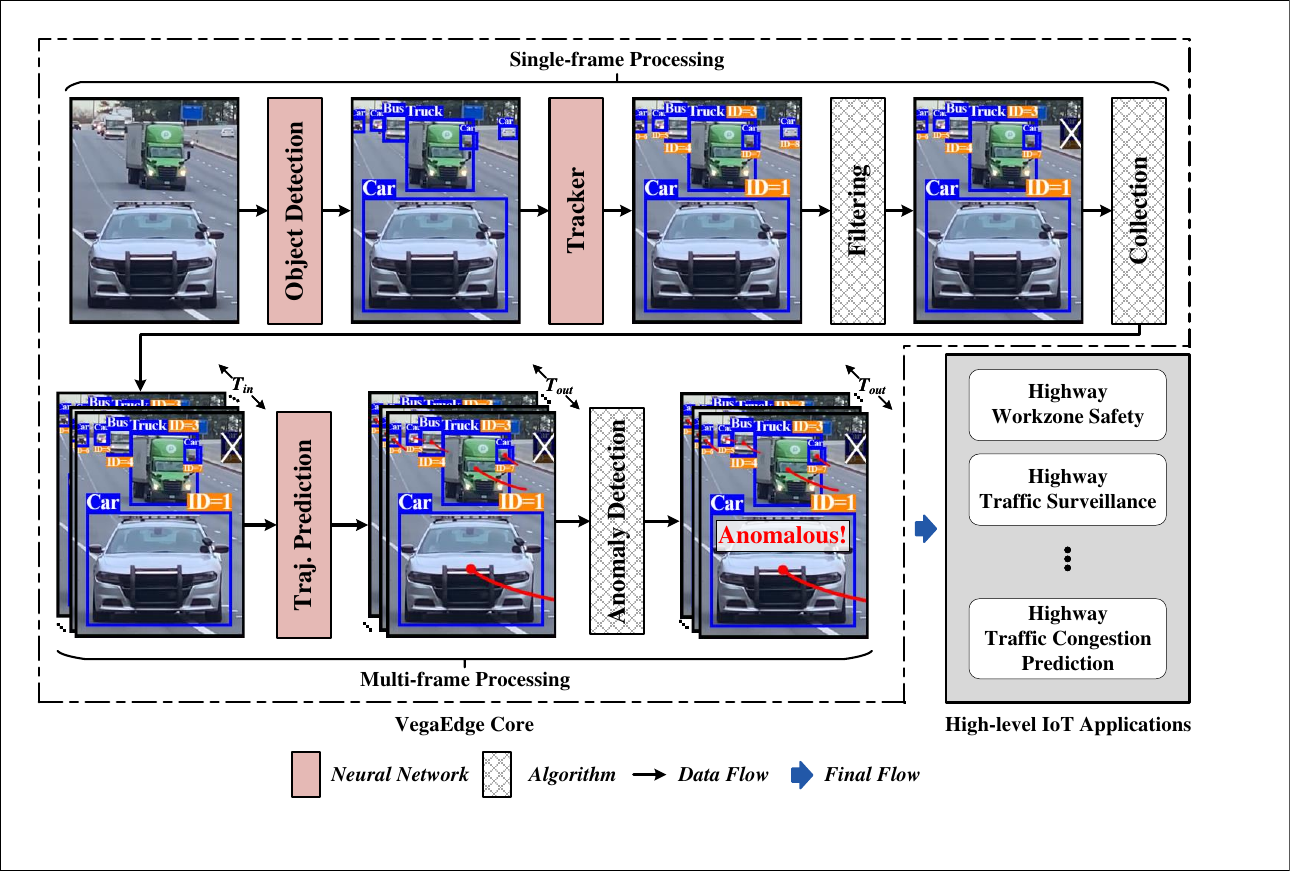}
    }
    \caption{Data flow and algorithmic design of VegaEdge. Object detector and tracking process the frames in a single-frame manner. The filtering algorithm ensures that the focus is on the correct side of the road, and the Collection algorithm collects the sufficient number of frames $T_{in}$ required by the Trajectory Prediction algorithm. The predicted path for $T_{out}$ frames in comparison with the actual path will be used for Anomaly Detection, which finally can be used for highway IoT applications at the edge.}
    \label{fig:main_diagram}
    \vspace*{-0.3cm}
\end{figure*}

\section{VegaEdge Design}

VegaEdge is an integration of high-performing AI models to empower IoT-embedded edge devices. Specifically designed to enhance real-time safety and surveillance on highways, its core capabilities encompass vehicle detection, tracking, and trajectory prediction, all converging toward the final goal of anomaly detection. Fig. \ref{fig:main_diagram} provides a step-by-step visual representation, illustrating how the entire VegaEdge system operates in a union. The high-level pseudocode of VegaEdge is also shown in Algorithm \ref{algo:vegaedge}.

As shown in Fig. \ref{fig:main_diagram} at a high level, VegeEdge detects vehicles within an image and subsequently tracks them across consecutive frames on a frame-by-frame basis. Following this, the system filters and accumulates the trajectories of distinct vehicles identified in the prior phase. Lastly, leveraging the gathered vehicle data from the past 3 seconds, it projects the trajectories for the upcoming five seconds, utilizing this foresight for effective anomaly detection. In the following subsections, we discuss our design choices and the working of each step shown in Fig. \ref{fig:main_diagram}.

\begin{algorithm}
\caption{ High-Level Workflow of VegaEdge }
\label{algo:vegaedge}
\begin{algorithmic}[1]
\REQUIRE{RGB image}
\ENSURE Detected anomalous vehicle trajectories

\STATE \textbf{Initialize:} Detection, tracking and prediction AI models
\STATE 3-sec Warmup for vehicle detection and ReID

\IF{frame is available}
    \STATE Detect vehicles in frames
    \STATE Assign unique IDs
    \STATE Ensure consistent IDs for previously identified vehicles 
\ELSE
\STATE Read the next frame
\ENDIF

\STATE Set \texttt{inference\_flag} to 0

\FOR{each unique vehicle ID in frame}
    \STATE Remove vehicles going away from the camera
    \STATE Remove IDs with less than 3 sec (15 frames) trajectory\IF{vehicles available for inference (3 sec trajectory)}
        \STATE Set \texttt{inference\_flag} to 1
    \ENDIF
\ENDFOR
\IF{\texttt{inference\_flag} == 1}
    \STATE \texttt{PishguVe} (Predict trajectory for next 5 seconds)
    \IF{trajectory available for anomaly}
        \FOR{ Each Vehicle ID available}
        \IF{Anomaly Criteria $>$ $T_{anomaly}$}
            \STATE Flag anomaly detected for specific vehicle ID
        \ENDIF
        \ENDFOR
    \ENDIF
\ENDIF
\STATE Go to step 3
\end{algorithmic}
\end{algorithm}

\subsection{Vehicle Detection}
For efficient and rapid vehicle detection for edge-integrated IoT devices, we opt for the YOLOv8l \cite{Jocher_YOLO_by_Ultralytics_2023} model and trained it on the BDD100k vehicle dataset \cite{yu2020bdd100k}. Our decision was influenced by the system's overall performance, such as latency, accuracy, and memory requirements.

Model size and performance are critical in edge deployments, particularly for IoT devices. YOLOv8l addresses this by being 35.9\% smaller than YOLOv8x \cite{Jocher_YOLO_by_Ultralytics_2023}, making it ideal for resource-limited embedded-IoT devices. Despite its reduced size, its mean Average Precision (mAP) is a competitive 52.9\%, only 1\% less than the 53.9\% of YOLOv8x.

\subsection{Vehicle Tracking}
Multi-object tracking (MOT) is fundamentally concerned with the identities of objects within video sequences. ByteTrack \cite{zhang2022bytetrack} uses an innovative association method that considers every detection box. Detection boxes with lower scores are processed by comparing their similarities with existing tracklets to accurately identify true objects while filtering out unwanted detections. 

Within this context, the VegaEdge employs the ByteTrack algorithm, renowned for its efficient and robust tracking capabilities. ByteTrack's architecture uses deep association techniques, ensuring consistent tracking across frames, even for challenges like occlusions and complex interactions. Its performance ia shown in Table \ref{tab:comparison_algo} on datasets like BDD100K and MOT20. Furthermore, ByteTrack boasts an impressive running speed without compromising on accuracy, making it an ideal choice for real-time applications such as VegaEdge. As discussed in \cite{zhang2022bytetrack}, ByteTrack achieves metrics of 80.3 MOTA, 77.3 IDF1, and 63.1 HOTA on the MOT17 test set, all while maintaining a 30 FPS running speed.

\subsection{Data Conditioning}
The output from detection and tracking is rigorously cleaned to ensure precise input, as the quality of the input directly dictates the accuracy of trajectory forecasts in the next step. To obtain precise vehicular trajectories, we've performed targeted data filtration and smoothing:
\begin{enumerate}
  \item Class-specific Inclusion: To eliminate potential noise from extraneous vehicular types, we selectively retain only cars, buses, and trucks.
  \item Unidirectional Movement: To focus on the flow of incoming vehicles, we exclude the vehicles operating in non-targeted directions, thereby standardizing the directional flow and reducing complexity.
  \item Temporal Presence Validation: Vehicles with transient appearances can introduce data anomalies. This validation process sets a minimum frame threshold, below which vehicular entries are deemed non-contributory and are subsequently removed.
  \item Trajectory Continuity: Despite thorough validation in previous steps, some trajectories may have missing frames. We fill such gaps through interpolation techniques, ensuring continuous and complete trajectories.
\end{enumerate}

In summary, the data cleansing and validation processes outlined above are crucial in ensuring the integrity and precision of vehicular trajectories used by VegaEdge's downstream tasks. By emphasizing class-specific inclusion, standardizing directional flow, validating temporal presence, and ensuring trajectory continuity, we lay the foundation for subsequent trajectory forecasts. This approach mitigates potential inaccuracies and fortifies our framework's reliability, positioning it to deliver consistent and high-quality results in real-world applications for which VegaEdge will be used.

\subsection{Trajectory Prediction}
We focus on highway-centric performance in the framework of VegaEdge's IoT applications. To meet this need, VegaEdge integrates the SotA PishguVe\cite{katariya2023pov} trajectory prediction algorithm on highway datasets, ensuring fast and accurate results without straining the device.

PishguVe was selected for its ability to make predictions at the pixel level, as shown in Table \ref{tab:comparison_algo}, its proven track record of state-of-the-art accuracy on multiple datasets \cite{NGSIM_US101, NGSIM_i80, katariya2023pov}, and its efficient memory footprint. Table \ref{tab:comparison_algo}, CHD-HA and CHD-EL represents CHD-High Angle and CHD-Eye-level trajectories from CHD dataset respectively.  Built on a fusion of graph isomorphism, convolutional neural networks, and attention mechanisms, PishguVe \cite{katariya2023pov} can forecast future vehicle positions with a model size of only 133K parameters. The input to PishguVe is past trajectories of vehicles are represented by a set of absolute coordinates, denoted as $P_{v}$, and a set of relative coordinates, denoted as $\Delta P_v$. The absolute coordinates are defined as $P_v = {(x_{v,p}^{t}, y_{v,p}^{t})}$, where ${t = 1, ..., T_{in}}$, ${v \in {1, 2, ..., n}}$ representing the index of the vehicle and ${x_{v,p}^{t}}$ and ${y_{v,p}^{t}}$ are x and y coordinates of the center of bounding box of vehicle ${v}$ at time ${t}$ for past trajectory denoted by $p$. The predicted future trajectories are shown as $\hat{F}_v = {(x_{v,f}^{t}, y_{v,f}^{t})}$, here ${t =  (T_{in}+1), ..., T_{pred}}$, $f$ denotes future trajectory, and ${v \in {1, 2, ..., n}}$, are generated as a set of coordinates for each vehicle in the image.

\subsection{Prediction-based Anomaly Detection}
VegaEdge uses the trajectory prediction-based anomaly detection approach discussed in section \ref{sec:Anomaly_Detection} of the paper, utilizing ADE and angle-based anomaly detection techniques. These methods offer a straightforward and efficient approach to detecting anomalies. This streamlined process makes our proposed method well-suited for integration within VegaEdge's IoT-based framework, which operates on hardware and time-constrained embedded IoT platforms. This efficiency allows for quick anomaly detection, enhancing the overall performance and responsiveness of the system. The performance of both approaches on two different datasets is demonstrated in the upcoming section.

\begin{table}[b]
\centering
\caption{Accuracy comparison of each algorithm.}
\label{tab:comparison_algo}
\resizebox{\columnwidth}{!}{%
\begin{tabular}{@{}cccc@{}}
\toprule
\textbf{Task}                             & \textbf{Method}                    & \textbf{Performance}                            & \textbf{Dataset}                    \\

\midrule
\multirow{3}{*}{Object Detection} & \multirow{3}{*}{YOLOv8l \cite{Jocher_YOLO_by_Ultralytics_2023}} & 52.9 (mAP)                         & COCO \cite{lin2014microsoft} \\
                                 &                           & \multicolumn{1}{l}{57.14 (mAP)@mAP50} & \multicolumn{1}{l}{BDD100K \cite{bdd100k}} \\
                                 &                           & 34.50 (mAP)@mAP50-95                              & BDD100K \cite{bdd100k}                      \\

\midrule
Tracking                         & ByteTrack \cite{zhang2022bytetrack}                 & 77.8 (MOTA)                            & MOT20 \cite{dendorfer2020mot20}                     \\ \midrule
\multirow{3}{*}{Path Prediction} & \multirow{3}{*}{PishguVe \cite{katariya2023pov}} & 20.75 (Pixels/ADE)                     & CHD-EL \cite{katariya2023pov} \\
                                 &                           & \multicolumn{1}{l}{16.81 (Pixels/ADE)} & \multicolumn{1}{l}{CHD-HA \cite{katariya2023pov}} \\
                                 &                           & 0.77(m/ADE)                            & NGSIM \cite{NGSIM_US101,NGSIM_i80}                       \\ \bottomrule
\end{tabular}%
}
\end{table}

\subsection{VegaEdge Performance Benchmarking}
\label{subsec:performance_bench}
VegaEdge's performance was evaluated across multiple platforms to understand its versatility and efficiency. Our testing platforms comprised a server with an Intel Xeon Silver 4114 CPU, Nvidia's V100 GPU, and Nvidia's Jetson Orin and Xavier NX boards. We chose the server setup as a reference point to contrast the performance metrics of the Jetson boards. These Jetson boards are designed for real-world tasks and are notable for their power efficiency, with Orin operating at 50W and Xavier NX at just 20W. Their efficiency and AI capabilities position them as ideal candidates for a wide range of IoT applications requiring edge computing.

Transitioning from the hardware evaluation, Table \ref{tab:comparison_algo} shows the performance metrics of three algorithms VegaEdge utilizes for crafting its workflow, as shown in Algorithm \ref{algo:vegaedge} and Figure \ref{fig:main_diagram}. In the domain of Object Detection, YOLOv8l achieves an mAP of 52.9 on COCO and 57.14 at mAP50 on BDD100K. It further scored 34.50 at mAP50-95. The Tracking algorithm, ByteTrack performs at a MOTA of 77.8 on the MOT20 dataset. Lastly, in trajectory prediction, the PishguVe algorithm was assessed on three distinctive datasets. It registered a Pixels/ADE of 20.75 and 16.81 on CHD-EL and CHD-HA, respectively, and a commendable m/ADE of 0.77 on NGSIM.

%% file: tex/results.tex
\section{Results}

\subsection{Anomaly detection on NGSIM Dataset:}
\label{subsec:ngsim_anomaly}
In this section, we evaluate our anomaly detection methodology on adversarial trajectories. These trajectories are derived using the NGSIM dataset with bird's eye camera-view of the trajectories, following the adversarial attack approach \cite{zhang2022adversarial}, adopted by \cite{jiao2023learning}. Our evaluation encompasses both the ADE and angle-based anomaly detection techniques. Our tests include both adversarial generated trajectories \cite{zhang2022adversarial} and actual real-world data from the NGSIM test dataset, similar to the study in \cite{alinezhad2023pishgu}.

\begin{figure}
    \centering
    \small
    \subfloat[\footnotesize{Fine grain (0.2s step)}]{%
        \includegraphics[width=0.48\linewidth]{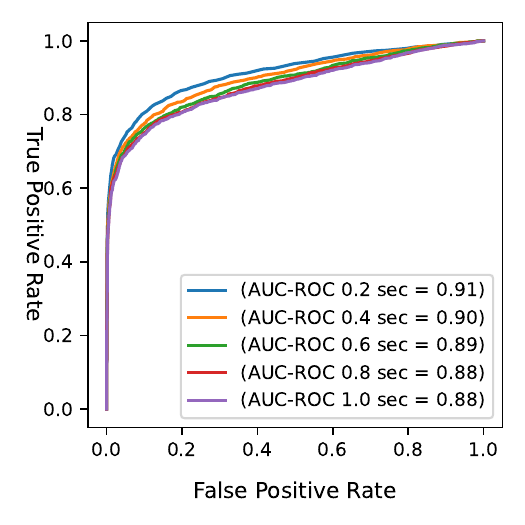}
        \label{fig:ngsim_ade_roc_0_2s}
    }
    \subfloat[\footnotesize{Coarse grain (1s step)}]{%
        \includegraphics[width=0.48\linewidth]{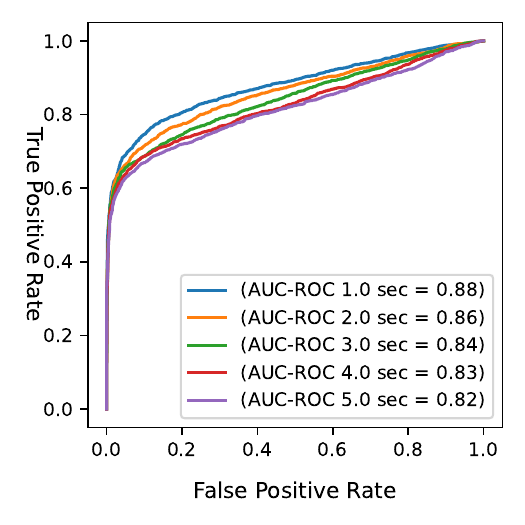}
        \label{fig:ngsim_ade_roc_1s}
    }
    
    \caption{AUC-ROC with varying detection windows for ADE-based anomaly detection method on NGSIM dataset. (a) Shows plot from 0s to 1s with time-step of 0.2 s (b) Shows plot from 1s to 5s with time-step of 1.0 s}
    \vspace*{-0.5cm}
    \label{fig:ngsim_ade_roc}
\end{figure}

\begin{figure}
    \centering
    \small
    \subfloat[\footnotesize{Fine grain (0.2s step)}]{%
        \includegraphics[width=0.48\linewidth]{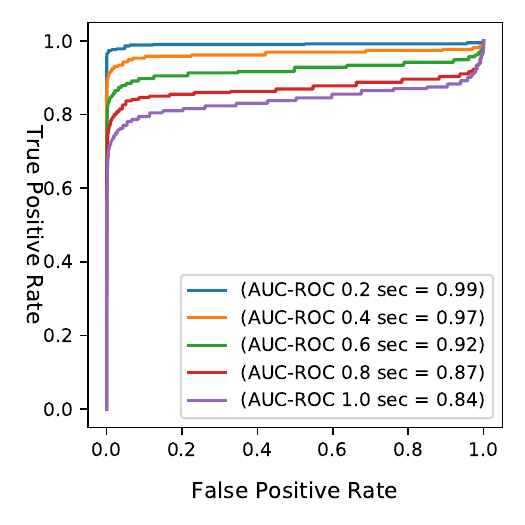}
        \label{fig:ngsim_angle_roc_0_2s}
    }
    \subfloat[\footnotesize{Coarse grain (1s step)}]{%
        \includegraphics[width=0.48\linewidth]{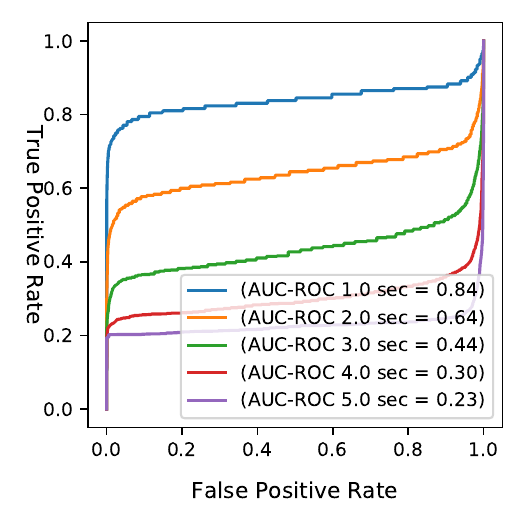}
        \label{fig:ngsim_angle_roc_1s}
    }
    \caption{AUC-ROC with varying detection windows for angle-based anomaly detection method on NGSIM dataset. (a) Shows plot from 0s to 1s with time-step of 0.2 s (b) Shows plot from 1s to 5s with time-step of 1.0 s}
    \vspace*{-0.3cm}
    \label{fig:ngsim_roc_angle}
\end{figure}

\begin{figure}
\vspace*{-0.3cm}
    \centering
    \small
    \subfloat[\footnotesize{ADE-based}]{%
        \includegraphics[width=0.48\linewidth]{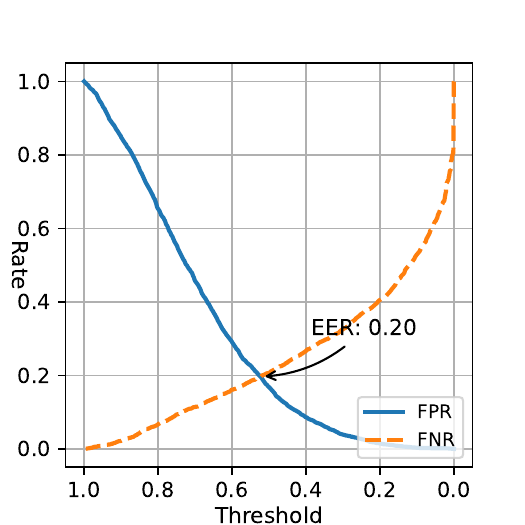}
        \label{fig:ngsim_eer_1s_ade}
    }
    \subfloat[\footnotesize{Angle-based}]{%
        \includegraphics[width=0.48\linewidth]{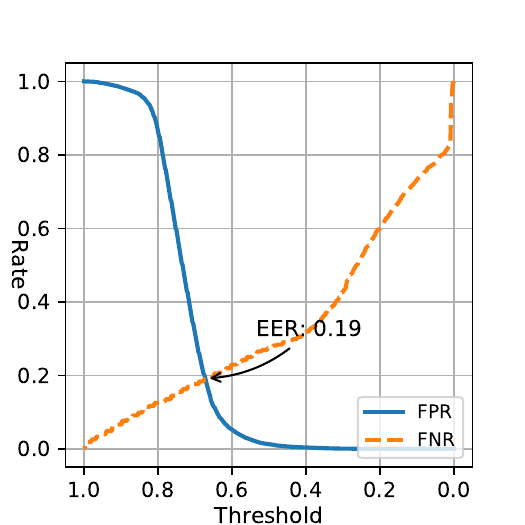}
        \label{fig:ngsim_eer_1s_angle}
    }
    \caption{EER plots for anomaly detections with 1s trajectory window for NGSIM dataset adversarial trajectories: (a) ADE-based anomaly, and (b) Angle-based anomaly.}
    \label{fig:ngsim_eer}
\end{figure}

\begin{table}[t]
    \centering
    \caption{EER for various Time Thresholds using ADE-and Angle based Anomaly on NGSIM Adversarial trajectories}
    \label{tab:ngsim_ade_angle_eer}
    \begin{tabular}{ccc}
    \toprule
        \multirow{2}{*}{\textbf{Time-step}} & \multicolumn{2}{c}{\textbf{EER}}\\
        \cmidrule(lr){2-3}
        \textbf{(s)} & ADE-based & Angle-based \\
        \midrule
        0.2 & 0.16 & 0.02\\
        0.4 & 0.17 & 0.05\\
        0.6 & 0.19 & 0.10\\
        0.8 & 0.19 & 0.15\\
        1.0 & 0.20 & 0.19\\
        2.0 & 0.22 & 0.38\\
        3.0 & 0.24 & 0.56\\
        4.0 & 0.25 & 0.69\\
        5.0 & 0.26 & 0.77\\
    \bottomrule
    \end{tabular}
    
\end{table}

The ADE-based anomaly detection method study revealed a distinct pattern regarding the Area Under the Receiver Operating Characteristic curve (AUC-ROC). 
The ADE-based anomaly detection method consistently performed best at an AUC-ROC of 0.91 for an ADE window of 0.2s, as visualized in Figs. \ref{fig:ngsim_ade_roc} and \ref{fig:ngsim_ade_roc_0_2s}. The angle-based approach initially outperformed the ADE-based method for short prediction windows, as seen in Fig. \ref{fig:ngsim_angle_roc_0_2s}, but its efficacy declined with larger windows, evident in Fig. \ref{fig:ngsim_angle_roc_1s}. Such behavior in the angle-based approach may stem from how anomalies are generated by applying constrained perturbations to real-world trajectories, making anomalies challenging to discern over more extended prediction periods.

The Equal Error Rate (EER) plot in Fig. \ref{fig:ngsim_eer} shows the EER value obtained by plotting False Negative and False Positive Rate for ADE and Angle anomaly methods for 1 second of predicted trajectory. Table \ref{tab:ngsim_ade_angle_eer} presents the EER for two anomaly detection methods on NGSIM adversarial trajectories across different time-step thresholds. The Angle method outperforms the ADE approach at shorter thresholds, such as 0.2s and 0.4s. However, as the time threshold grows, their performance converges. By 5.0 seconds, the EERs are 0.26 for ADE and 0.77 for the Angle method, indicating a faster performance drop for the Angle approach over extended time steps. 

\subsection{Anomaly detection on CAD}
\label{subsec:anomaly_cad}
\begin{figure}[!t]
    \centering
    \small
    \subfloat[\footnotesize{Fine grain (0.2s)}]{%
        \includegraphics[width=0.48\linewidth]{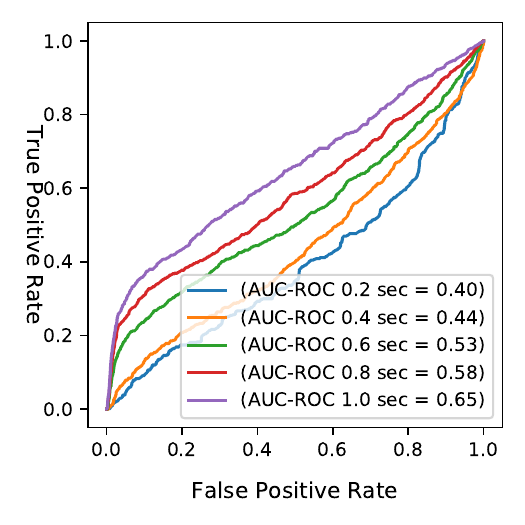}
        \label{fig:chd_ade_roc_0_2s}
    }
    \hfill  
    \subfloat[\footnotesize{Coarse grain (1s)}]{%
        \includegraphics[width=0.48\linewidth]{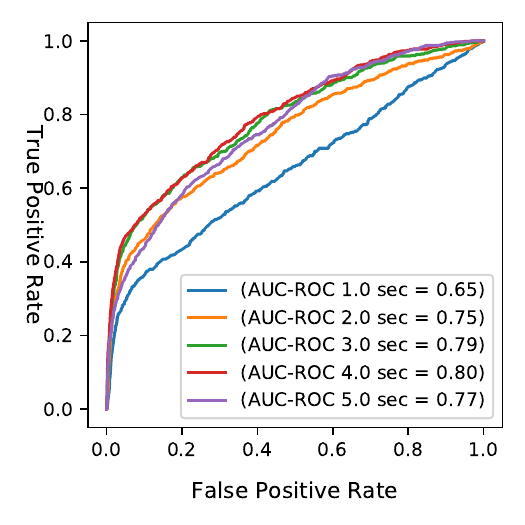}
        \label{fig:chd_ade_roc_1s}
    }
    \caption{AUC-ROC with varying prediction windows for ADE-based anomaly detection method on CAD. (a) Shows plot from 0s to 1s with a time-step of 0.2 s. (b) Shows plot from 1s to 5s with time-step of 1.0 s.}
    \vspace*{-0.5cm}
    \label{fig:chd_ade_roc}
\end{figure}

\begin{figure}[!t]
    \centering
    \small
    \subfloat[\footnotesize{Fine grain (0.2s)}]{%
        \includegraphics[width=0.48\linewidth]{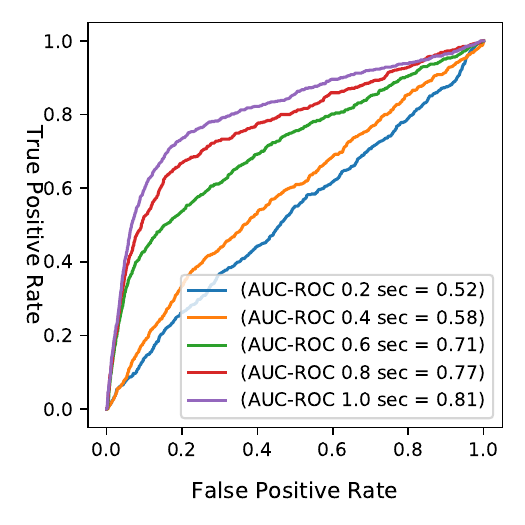}
        \label{fig:CHD_angle_roc_0_2s}
    }
    \hfill  
    \subfloat[\footnotesize{Coarse grain (1s)}]{%
        \includegraphics[width=0.48\linewidth]{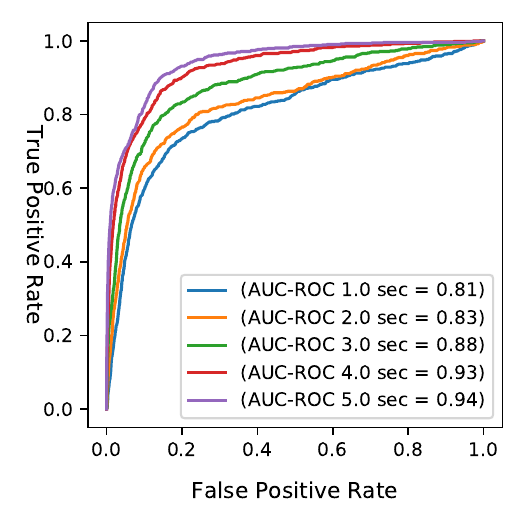}
        \label{fig:CHD_angle_roc_1s}
    }
    \caption{AUC-ROC with varying prediction windows for Angle-based anomaly detection method on CAD. (a) Shows plot from 0s to 1s with a time-step of 0.2 s. (b) Shows plot from 1s to 5s with time-step of 1.0 s.}
    \vspace*{-0.5cm}
    \label{fig:chd_angle_roc}
\end{figure}

In this section, we evaluate our anomaly detection approach on real-world trajectories sourced from CAD consisting of high-angle (CHD-HA) and eye-level (AHD-EL) camera-view of highway vehicles, as introduced in section \ref{sec:CAD}. To offer a comprehensive view of the results, the AUC-ROC values for both fine and coarse grain time steps are graphically represented in Fig. \ref{fig:chd_ade_roc} for the ADE-based anomaly detection method and in Fig. \ref{fig:chd_angle_roc} for the angle-based method.

Expanding on the earlier analysis, Table \ref{tab:chd_ade_angle_eer} provides a breakdown of the Equal Error Rate (EER) performance for various time thresholds, contrasting the ADE and angle-based anomaly detection methods on CAD's data. It is clear that the Angle approach consistently outperforms the ADE method across all examined time steps. Starting from an EER of 0.48 at the 0.2s mark, the angle method demonstrates a steady improvement, with an EER of 0.12 at the 5s threshold. In contrast, the ADE-based method initiates with an EER of 0.58 at 0.2s and gradually improves to 0.32 at 5s. These metrics show the superior efficacy of the angle-based approach, especially since the prediction windows are elongated, making it a more robust choice for anomaly detection in the context of the CAD.

\subsection{Real-world vs Adversarial Anomaly Trajectories}
Intriguingly, the real-world trajectories demonstrate an inverse trend compared to the results of the adversarial anomaly dataset. Specifically, the AUC-ROC values for the ADE anomaly detection method peak at a 4 and 5-second prediction window, recording the area of 0.80 for a 4s window as shown in Fig. \ref{fig:chd_ade_roc_1s}. This suggests a higher sensitivity of the ADE method to longer prediction windows when applied to real-world data. Similarly, in Fig. \ref{fig:CHD_angle_roc_1s} the angle-based method exhibits stellar performance with an AUC-ROC of 0.94 for the same 5s window. Such observations indicate that while synthetic or constrained trajectory datasets may favor short prediction windows, real-world trajectories might inherently contain more distinguishable anomalies in longer prediction intervals. 

Comparing the EER values from the real-world CAD trajectories (Table \ref{tab:chd_ade_angle_eer}) with those from the adversarial NGSIM trajectories (Table \ref{tab:ngsim_ade_angle_eer}), several striking differences emerge. The NGSIM adversarial trajectories exhibit substantially lower EER values in the initial time steps, especially for the angle-based approach. For example, at the 0.2-second mark, the NGSIM data record a notably lower EER of 0.02 for the angle-based method than 0.48 for the CAD dataset.

\begin{figure}[!t]
\vspace*{-0.3cm}
    \centering
    \small
    \subfloat[\footnotesize{ADE-based}]{%
        \includegraphics[width=0.48\linewidth]{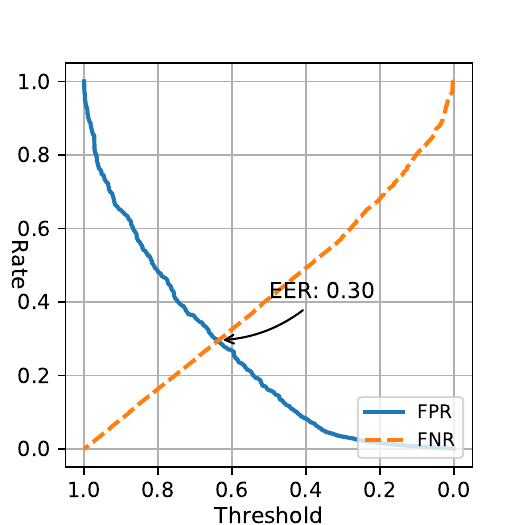}
        \label{fig:cad_eer_1s_ade}
    }
    \hfill  
    \subfloat[\footnotesize{Angle-based}]{%
        \includegraphics[width=0.48\linewidth]{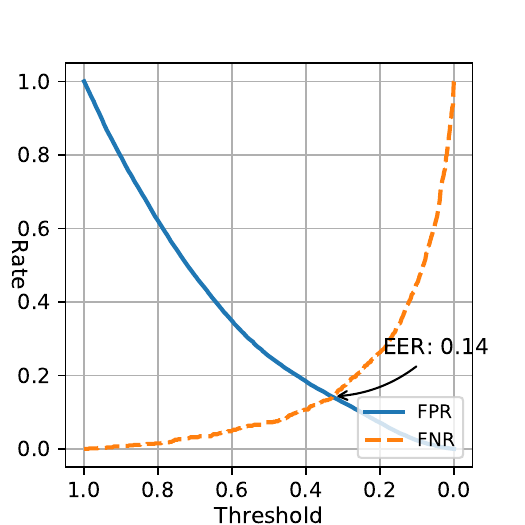}
        \label{fig:cad_eer_1s_angle}
    }
    \caption{EER plots for 4s anomaly detections for CAD dataset: (a) ADE-based anomaly, and (b) Angle-based anomaly.}
    \label{fig:cad_eer}
\end{figure}

\begin{table}[t]
    \centering
    \caption{EER for various Time Thresholds using ADE and Angle based Anomaly on CAD}
    \label{tab:chd_ade_angle_eer}
    \begin{tabular}{ccc}
    \toprule
        \multirow{2}{*}{\textbf{Time-step}} & \multicolumn{2}{c}{\textbf{EER}}\\
        \cmidrule(lr){2-3}
        \textbf{(s)} & \textbf{ADE-based} & \textbf{Angle-based} \\
        \midrule
        0.2& 0.58 & 0.48 \\
        0.4 & 0.55 & 0.43\\
        0.6 & 0.50 & 0.35\\
        0.8 & 0.46 & 0.28\\
        1.0 & 0.41 & 0.24\\
        2.0 & 0.34 & 0.22\\
        3.0 & 0.30 & 0.17\\
        4.0 & 0.30 & 0.14\\
        5.0 & 0.32 & 0.12\\
    \bottomrule
    \end{tabular}
    
\end{table}

A potential reason for this marked divergence might be the inherent nature of the datasets. The NGSIM adversarial trajectories, being synthetically generated, likely present more pronounced and discernible anomalies that the detection methods can more readily identify, especially within shorter prediction windows. In contrast, the CAD real-world trajectories, a genuine reflection of real-world driving behaviors, might be embedded with subtler and more intricate anomalies. These nuances could pose more significant challenges in detection, resulting in higher EER values, especially in the shorter time-step intervals. Moreover, the complexities and variances found in real-world driving behaviors could introduce a wider array of anomalies, making distinguishing between normal and anomalous patterns more intricate for the detection methods when applied to the CAD.


\subsection{VegaEdge Performance}
\input{tex/VegaEdge_performance}




\begin{table}
\centering
\caption{Throughput (processed trajectories per second) comparison of VegaEdge on the three platforms across different road types and traffic densities (expressed as vehicle trajectories processed per second) }
\label{tab:latency_and_fps}
\begin{tabular}{cccc}
\toprule
\multirow{2}{*}{\textbf{Road Type}}& \multicolumn{3}{c}{\textbf{Throughput}} \\ 
\cmidrule(lr){2-4} 
 \textbf{(Traffic Density)} & \textbf{Server} & \textbf{Jetson Orin} & \textbf{Xaview NX} \\
\midrule
 3 lanes and merger (140) & 1770 & 758 & 243 \\
2 lanes with workzone (18) & 2868 & 132 & 47 \\
2 lanes (simulated video) (13) & 1050 & 92 & 31 \\

\bottomrule
\end{tabular}
\vspace{0.2cm}
\end{table}

\begin{table}[t]
    \centering
    \caption{Comparison of Anomaly detection Prediction Windows, Buffer Times (Excluding 3s Input Trajectory), and Vehicle Distances from Camera at 60 mph.}
    \label{tab:vegaedge_buffertimes}
    \begin{tabular}{ccc}
    \toprule
        \textbf{Error Detection} & \textbf{Buffer} & \textbf{Distance}\\
        \textbf{Window (s)} & \textbf{Time (s)} & \textbf{(m)} \\
        \midrule
        0.2 & 8.8 & 235\\
        1.0 & 8 & 213\\
        3.0 & 6 & 160\\
        5.0 & 4 & 107\\        
    \bottomrule
    \end{tabular}
\end{table}

\subsection{Highway Workzone Safety Application}

In highway work zones, safeguarding workers from oncoming vehicles is a primary concern. Through trajectory prediction combined with anomaly detection, VegaEdge detects trajectories that may pose threats and alerts workers. To achieve this objective, the proposed design must demonstrate real-time performance on edge devices. The efficacy of VegaEdge, particularly when implemented on Jetson Orin, is exemplified in Table \ref{tab:vegaedge_buffertimes}.

Given an error detection window of 0.2s, VegaEdge provides a buffer time of 8.8s, corresponding to a 235m distance from a camera for a vehicle moving at 60 mph. As the detection window widens, the buffer diminishes but remains noteworthy, with 1s, 3s, and 5s windows yielding buffer times of 8s, 6s, and 4s, respectively. A clear trade-off emerges between larger buffer windows and heightened accuracy, as noted in the anomaly detection results for CAD in section \ref{subsec:anomaly_cad}.

In hazardous highway work zones, these buffer times are beneficial and vital. Even minor increments in warning time can significantly alter outcomes. VegaEdge's ability to grant these buffers, particularly on Jetson platforms, underscores its practicality in real-world scenarios and its role in bolstering worker safety.

\begin{table}[b]
\centering
\caption{Comparison of VegaEdge Power consumption and Throughput ((Trajectories processed per second) on Jetson Orin Power Modes for real word highway ($\sim140$  Vehicles detected per second across 30 frames.). Power is measured in Watts.}
\label{tab:power_comparison}
\begin{tabular}{ccccc}
\toprule
\multirow{2}{*}{\textbf{Power}} & \multirow{2}{*}{\textbf{Total}} & \multirow{2}{*}{\textbf{GPU} }& \multirow{2}{*}{\textbf{CPU}} & \multirow{2}{*}{\textbf{Throughput}} \\
\\[-1.0ex]
{\textbf{Mode}} & \textbf{Power} & \textbf{Power} & \textbf{Power} &  \\

\midrule
MAXN & 18.14 & 3.66 & 8.80 & 758 \\
30W  & 11.44 & 3.09 & 3.72  & 477 \\
15W  & 8.43 & 2.82 & 1.3  & 214 \\
\bottomrule
\end{tabular}
\vspace{0.2cm}
\end{table}

\subsection{VegaEdge Power Analysis on IoT Platform}
In this subsection, we conduct a comprehensive power consumption analysis of the Nvidia Jetson Orin platform with the primary objective of assessing the practical utility and performance of VegaEdge across various power modes. We report the total power consumed across all the channels ($P_{Total}$) calculated using the following equation:
\begin{equation}
P_{\text{Total}} = P_{\text{GPU}} + P_{\text{CPU}} + P_{\text{IOs}} + P_{\text{AO}}
\end{equation}

Where $P_{GPU}$ is the total power consumed by the GPU and SOC cores, $P_{CPU}$ is the power consumed by the CPU and CV cores, and $P_{\text{IOs}}$ is the power consumed by the system's 5V rail for various input and output ports, respectively on one of the power monitors \cite{nvidia2023power}. $P_{\text{AO}}$ stands for power consumed by Always On (AO) power rail on another power monitor \cite{nvidia2023power}.

In Table \ref{tab:power_comparison}, VegaEdge's power consumption and throughput on the Jetson Orin are evaluated across different power modes for a real-world highway scenario. At the MAXN (40W) setting, the system processes 758 trajectories per second, consuming 18.14W, and reducing the power mode to 30W and 15W results in decreased throughputs of 477 and 214 trajectories per second, respectively, with corresponding reductions in power consumption. Despite the higher power demands compared to typical IoT devices, VegaEdge on Jetson Orin showcases a valuable trade-off between power consumption and high-throughput processing, making it a viable solution for edge applications requiring rapid data processing.

%% file: tex/VegaEdge_performance.tex
In our evaluation, we primarily focus on the performance of VegaEdge on the Jetson platforms. With its superior computational capabilities, the server platform serves as a benchmarking reference to underscore the efficiency and feasibility of deploying VegaEdge in more constrained environments.

\subsubsection{Performance on Jetson Platforms}

Table \ref{tab:latency_and_fps} delineates the throughput of VegaEdge across different road scenarios and traffic densities. On Jetson platforms, VegaEdge's performance showcases its adaptability and efficiency, particularly given the embedded nature of these devices.

For the \textit{3 lanes high traffic density} scenario, the Jetson Orin processes 758 trajectories every second, with 140 unique vehicles detected per second and the intricate nature of merging scenarios. This throughput ensures real-time processing capabilities essential for many applications. Meanwhile, the Xaview NX, while trailing with 243 trajectories per second, still provides a usable metric for less time-sensitive tasks or preliminary data-gathering efforts.

The \textit{2 lanes with workzone with low traffic density} scenario provides a different challenge, simulating a common urban environment with traffic modulations due to work zones. In this context, the Jetson Orin delivers a performance of 132 trajectories per second, making it ideal for surveillance and alert systems in smart city setups. This reduced throughput is attributed to the fewer vehicles present in the scene, not the capability constraints. Meanwhile, the Xavier NX offers 47 trajectories per second, which may be apt for tasks requiring less frequent monitoring.

\subsubsection{Digital Twin Systems}

Another aspect of our evaluation is the \textit{2 lanes (simulated video) (13)} scenario. The Jetson Orin's capability to process 92 trajectories per second in a simulated environment underscores its potential utility in digital twin systems. Digital twin systems, which mirror and simulate real-world environments, are important in predictive maintenance, system optimization, and various simulation-driven tasks. The ability of VegaEdge to run efficiently on simulated data on the Jetson Orin emphasizes its versatility and readiness for such advanced applications.